\documentclass[10pt,onecolumn,twoside, english,a4paper]{article}

\usepackage[T1]{fontenc}
\usepackage[latin1]{inputenc}
\usepackage{epsf,epsfig,subfigure,latexsym,latexsym,amssymb,alltt}
\usepackage{url}
\usepackage{babel}
\usepackage{balance}
\usepackage{cite}
\usepackage{multirow}
\usepackage{geometry}
\usepackage{float}
\restylefloat{table}

\def\keywords#1{\noindent\textbf{Keywords.} #1}
\def\abstract#1{\noindent\textbf{Abstract.} \textit{#1}}
\def\authorblock[#1]#2{\textbf{\normalsize #1} \\#2}
\def\aff#1{\normalsize #1}
\def\dept#1{\normalsize #1}
\def\address#1{\normalsize #1}
\def\email#1{\texttt{\normalsize #1}}

\title{\textbf{Preliminary Report on the Structure of Croatian Linguistic Co-occurrence Networks}}
\author{
\authorblock[Domagoj Margan, Sanda Martin\v{c}i\'c-Ip\v{s}i\'c, Ana Me\v{s}trovi\'c]{
\aff{Department of Informatics}\\
\dept{University of Rijeka}\\
\address{Radmile Matej\v{c}i\'c 2, 51000 Rijeka, Croatia}\\
\email{\{dmargan, smarti, amestrovic\}@uniri.hr}\\
}}
\date{  }

\begin{document}
\maketitle
\thispagestyle{empty}
\abstract{In this article, we investigate the structure of Croatian linguistic co-occurrence
networks. We examine the change of network structure properties by systematically varying the
co-occurrence window sizes, the corpus sizes and removing stopwords. In a co-occurrence window of size $n$ we establish a link between the current word and $n-1$ subsequent words. The results point out that the increase of the co-occurrence window size is followed by a decrease in diameter, average path shortening and expectedly condensing the average clustering coefficient. The same can be noticed for the removal of the stopwords. Finally, since the size of texts is reflected in the network properties, our results suggest that the corpus influence can be reduced by increasing the co-occurrence window size.\\}


\section{Introduction}
The complex networks sub-discipline tasked with the analysis of language has been recently associated with the term of linguistic's network analysis. Text can be represented as a complex network of linked words: each individual word is a node and interactions amongst words are links. The interactions can be derived at different levels: structure, semantics, dependencies, etc. Commonly they rise from a simple criterion such as co-occurrence of two words within a sentence, or text. 
 
The pioneering construction of linguistic networks was in 2001, when Ferrer i Cancho and Sol\'e
\cite{i2001small} showed that the co-occurrence network from the British National Corpus has a small average path length, a high clustering  coefficient, and a two-regime power law degree distribution; the
network exhibits small-world and scale-free properties. Drogotsev and Mendes
\cite{dorogovtsev2001language} used complex networks to study language as a self-organizing network
of interacting words. The co-occurrence networks were constructed by linking two neighboring words
within a sentence. Masucci and Rodgers \cite{masucci2006network} investigated the network topology of
Orwell's `1984' focusing on the local properties: nearest neighbors and the clustering coefficient by
linking the neighboring words. Pardo \textit{et al.} \cite{pardo2006using} used the complex network's
clustering coefficient as the measure of text summarization performance. The original and summarized
texts were preprocessed with stopwords' removal and lemmatization. For the network construction they
used reversed window orientation which caused the word to be connected to the previous words with forwarding links' directions. Caldiera \textit{et al.}
\cite{caldeira2006network} examined the structure of the texts of individual authors. After stopword
elimination and lemmatization each sentence was added to the network as a clique\footnote{A clique in
an undirected network is a subset of its nodes such that every two nodes in the subset are linked.}.
Biemann \textit{et al.} \cite{biemann2012quantifying} compared networks where two neighboring words
were linked with networks where all the words co-occurring in the sentence were linked. From the network properties they
derived a quantifiable measure of generative language (n-gram artificial language) regarding the
semantics of natural language. Borge-Holthoefer \cite{borge2010semantic} produced a methodological and
formal overview of complex networks from the language research perspective. Liu and Cong
\cite{liu2013language} used complex network parameters for the classification (hierarchical clustering)
of 14 languages, where Croatian was amongst 12 Slavic.

In this paper we construct the linguistic co-occurrence networks from Croatian texts. We examine
the change of a network's structure properties by systematically varying the co-occurrence window sizes,
the corpus sizes and stopwords' removal. In a co-occurrence window of size $n$ we establish
a link between the current word and $n-1$ subsequent words.

In Section 2 we define network properties needed to accurately analyze small-world and scale-free
characteristics of co-occurrence networks, such as diameter, average path length and average
clustering coefficient. In Section 3 we present the construction of 30 co-occurrence networks. The network measurements are in Section 4. In the final Section, we elaborate on the obtained results
and make conclusions regarding future work.

\section{The network structure analysis}
In the network $N$ is the number of nodes and  $K$ is the number of links. In weighted networks every link
connecting two nodes has an associated weight $w \in {R}_{0}^{+}$. The co-occurrence window $m_{n}$ of
size $n$ is defined as $n$ subsequent words from a text. The number of network components is denoted
by $\omega$. 

For every two connected nodes $i$ and $j$ the number of links lying on the shortest path between them is
denoted as $d_{ij}$, therefore the average distance of a node $i$ from all other nodes is:
\begin{equation}
d_{i} = \frac{\sum_{j}d_{ij}}{N}.
\end{equation} 

And the average path length between every two nodes $i,j$ is:
\begin{equation}
L = \sum_{i,j} \frac{d_{ij}}{N(N-1)}.
\end{equation}

The maximum distance results in the network diameter:
\begin{equation}
D = max_{i} d_{i}.
\end{equation}

For weighted networks the clustering coefficient of a node $i$ is defined as the geometric average of
the subgraph link weights:
\begin{equation}
c_{i} = \frac{1}{k_{i}(k_{i}-1)} \sum_{ij} (\hat{w}_{ij} \hat{w}_{ik} \hat{w}_{jk})^{1/3},
\end{equation}

where the link weights $\hat{w}_{ij}$ are normalized by the maximum weight in the network
$\hat{w}_{ij} = w_{ij}/\max(w)$. The value of $c_{i}$ is assigned to 0 if $k_{i} < 2$. 

The average clustering of a network is defined as the average value of the clustering coefficients of
all nodes in a network:
\begin{equation}
C = \frac{1}{N}\sum_{i} c_{i}.
\end{equation} 

If $\omega > 1$, $C$ is computed for the largest network component.

An important property of complex networks is degree distribution. For many real networks this
distribution follows power law \cite{newman2005power}, which is defined as:
\begin{equation}
P(k) \sim k^{- \alpha}.
\end{equation}

\section{Network construction}
\subsection{Data}
For the construction and analysis of co-occurrence networks, we used a corpus of literature, containing 10
books written in or translated into the Croatian language. For the experiments we divided the corpus into three
parts: C1 - one book, C2 - four books and C3 - ten books, where C1 $\subseteq$ C2 $\subseteq$ C3, as shown in Table 1. 

Stopwords are a list of the most common, short function words which do not carry strong semantic properties, 
but are needed for the syntax of 
language (pronouns, prepositions, conjunctions, abbreviations, interjections,...). The Croatian
stopwords list contains 2,923 words in their inflected forms. Examples of stopwords are: `is', `but',
`and', `which', `on', `any', `some'.
 
\begin{table}[h!]
\caption{The statistics for the corpus of 10 books}
\begin{center}
\begin{tabular}{||l|c|c|c||}
\hline
Corpus part & C1 & C2 & C3 \\ \hline
\# of words & 28671 & 252328 & 895547 \\ \hline
\# of unique words & 9159 & 40221 & 91018 \\ \hline
\# of stopwords & 371 & 588 & 629 \\
\hline
\end{tabular}
\end{center}
\end{table}

\subsection{The construction of co-occurrence networks}
We constructed 30 different co-occurrence networks, weighted and directed, from the corpus in Table 1.
Words are nodes, and they are linked if they are in the same sentence according to the size of the
co-occurrence window. The co-occurrence window $m_{n}$ of size $n$ is defined as a set of $n$ subsequent words
from a text. Within a window the links are established between the first word and $n-1$ subsequent
words. During the construction we considered the sentence boundary as the window boundary too. Three
steps in the network construction for a sentence of 5 words, and the co-occurrence window size
$n=2..5$ is shown in Fig. 1. 

\begin{figure}[h!]
\centering 
\includegraphics[width=0.7\textwidth]{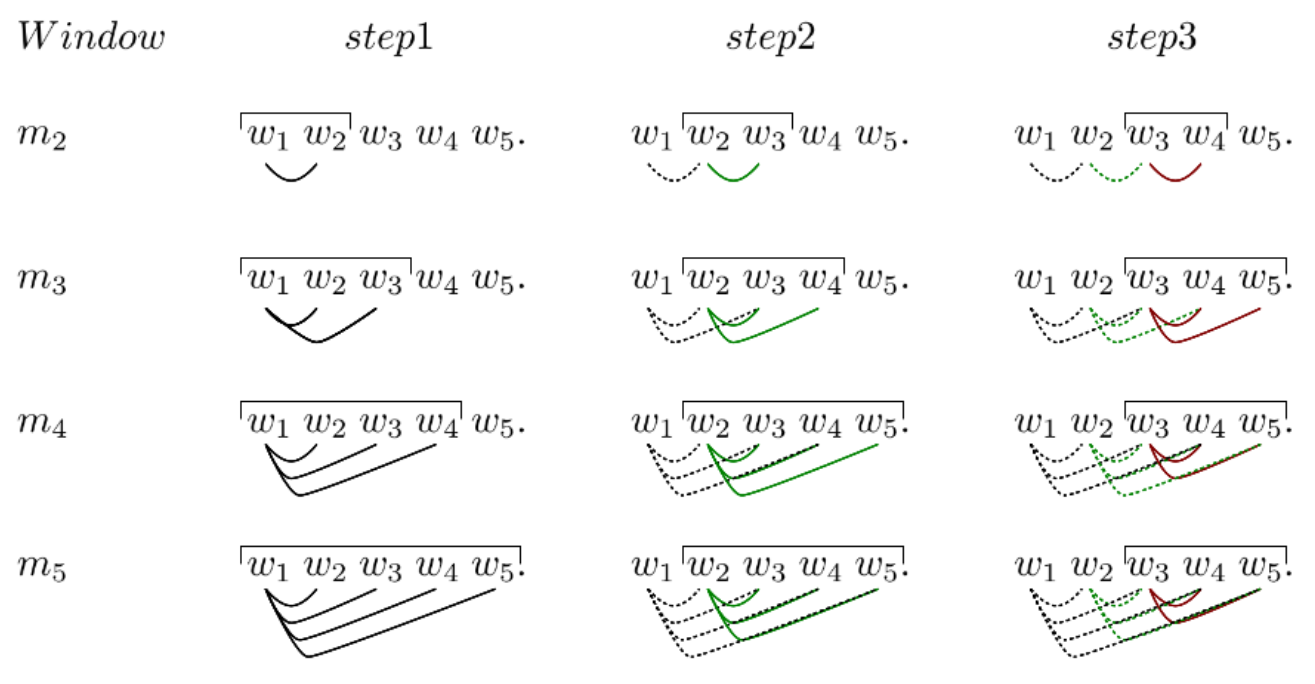}
\caption{An illustration of 3 steps in a network construction with a co-occurrence window $m_{n}$ of
sizes $n=2...5$. $w_{1}...w_{5}$ are words within a sentence.}
\end{figure}

The weight of the link between two nodes is proportional to the overall co-occurrence frequencies of the corresponding words within a co-occurrence window. For all three parts of the corpus C1, C2, C3, we
examined the properties of co-occurrence networks constructed with various $m_{n}$, $n = 2,3,4,5,6$.
Besides 5 window sizes for co-occurrence networks, we also differentiate upon the criterion of the inclusion or exclusion of stopwords.  

Network construction and analysis was implemented with the Python programming language using the
NetworkX software package developed for the creation, manipulation, and study of the structure,
dynamics, and functions of complex networks \cite{hagberg2008exploring}. Numerical analysis and
visualization of power law distributions was made with the `powerlaw' software package
\cite{alstott2013powerlaw} for the Python programming language.

\section{Results}
\begin{table}[h!]
\caption{Networks constructed from C1. Measures noted with the $sw$ subscript are results with stopwords included.}
\begin{center}
\begin{tabular}{||l|c|c|c|c|c||}
\hline & $m_{2}$ & $m_{3}$ & $m_{4}$ & $m_{5}$ & $m_{6}$ \\ \hline\hline
$N_{sw}$ & 9530 & 9530 & 9530 & 9530 & 9530 \\ \hline $N$ & 9159 & 9159 & 9159 & 9159 & 9159 \\
\hline\hline 
$K_{sw}$ & 22305 & 43894 & 64161 & 83192 & 101104 \\ \hline $K$ & 14627 & 28494 & 41472 & 53596 & 64840 \\ \hline\hline 
$L_{sw}$ & 3.59 & 2.92 & 2.70 & 2.55 & 2.45 \\ \hline $L$ & 6.42 & 4.73 & 4.12 & 3.79 & 3.58 \\
\hline\hline 
$D_{sw}$ & 16 & 9 & 7 & 6 & 6 \\ \hline $D$ & 26 & 15 & 11 & 10 & 8\\ \hline\hline 
$C_{sw}$ & 0.15 & 0.55 & 0.63 & 0.66 & 0.68 \\ \hline $C$ & 0.01 & 0.47 & 0.56 & 0.60 & 0.64 \\
\hline
$\omega_{sw}$ & 5 & 5 & 5 & 5 & 5 \\ \hline $\omega$ & 15 & 15 & 15 & 15 & 15 \\ \hline\hline 
\end{tabular}
\end{center}
\end{table}

The comparisons of the properties for networks differing in the co-occurrence window size are shown in Tables
2, 3 and 4. Clearly, the results show that the networks constructed with larger co-occurrence window
emphasize small-world properties. More precisely, the values of the average path length and network
diameter decrease proportionally to the increase of co-occurrence window size. Likewise, the
average clustering coefficient becomes larger in accordance with the increment of $m_{n}$. 

\begin{table}[h!]
\caption{Networks constructed from C2. Measures noted with the $sw$ subscript are results with stopwords included.}
\begin{center}
\begin{tabular}{||l|c|c|c|c|c||}
\hline & $m_{2}$ & $m_{3}$ & $m_{4}$ & $m_{5}$ & $m_{6}$ \\ \hline\hline
$N_{sw}$ & 40809 & 40809 & 40809 & 40809 & 40809  \\ \hline $N$ & 40221 & 40221 & 40221 & 40221 & 40221 \\ \hline\hline
$K_{sw}$ & 156857 & 307633 & 445812 & 572463 & 688484 \\ \hline $K$ & 108449 & 207437 & 296233 &
375535 & 446547 \\ \hline\hline
$L_{sw}$ & 3.25 & 2.81 & 2.64 & 2.52 & 2.43  \\ \hline $L$ & 4.69 & 3.86 & 3.54 & 3.35 & 3.23 \\
\hline\hline
$D_{sw}$ & 18 & 12 & 8 & 7 & 6 \\ \hline $D$ & 24 & 14 & 11 & 9 & 9 \\ \hline\hline
$C_{sw}$ & 0.25 & 0.58 & 0.65 & 0.68 & 0.70  \\ \hline $C$ & 0.02 & 0.43 & 0.52 & 0.56 & 0.59 \\ 
\hline\hline
$\omega_{sw}$ & 9 & 9 & 9 & 9 & 9 \\ \hline $\omega$ & 33 & 33 & 33 & 33 & 33 \\  \hline
\end{tabular}
\end{center}
\end{table}

\begin{table}[h!]
\caption{Networks constructed from C3. Measures noted with the $sw$ subscript are results with stopwords included.}
\begin{center}
\begin{tabular}{||l|c|c|c|c|c||}
\hline
& $m_{2}$ & $m_{3}$ & $m_{4}$ & $m_{5}$ & $m_{6}$ \\ \hline 
$N_{sw}$ & 91647 & 91647 & 91647 & 91647 & 91647  \\ \hline $N$ & 91018 & 91018 & 91018 & 91018 & 91018 \\ \hline\hline 
$K_{sw}$ & 464029 & 911277 & 1315888 & 1680848 & 2009187 \\ \hline $K$ & 360653 & 684008 & 963078 & 1202869 & 1409599\\ \hline\hline 
$L_{sw}$ & 3.10 & 2.74 & 2.58 & 2.47 & 2.38 \\ \hline $L$ & 4.17 & 3.55 & 3.30 & 3.16 & 3.08 \\
\hline\hline 
$D_{sw}$ & 23 & 13 & 9 & 7 & 7 \\ \hline $D$ & 34 & 19 & 14 & 11 & 9 \\ \hline\hline 
$C_{sw}$ & 0.32 & 0.61 & 0.67 & 0.69 & 0.71 \\ \hline $C$ & 0.03 & 0.42 & 0.51 & 0.55 & 0.58\\
\hline\hline
$\omega_{sw}$ & 22 & 22 & 22 & 22 & 22 \\ \hline $\omega$ & 64 & 64 & 64 & 64 & 64 \\ \hline 
\end{tabular}
\end{center}
\end{table}

In Tables 2, 3 and 4 we also compare the characteristics of networks with the removal of the stopwords. 
In addition to the proportional strengthening of small-world
properties with the increase of $m_{n}$, the same phenomenon appears with the inclusion of stopwords
in the process of building the network. All of the networks show smaller network distance measures
and greater clustering coefficient with the stopwords included. 

Furthermore, stopwords have an impact on the average clustering coefficient in a way that increasing the
corpus size with the stopwords included will result in a higher clustering coefficient, while increasing
the corpus size with the stopwords excluded will result in a lower clustering coefficient (Fig. 2).
This may be explained by the high impact of stopwords as the main hubs. Table 5 shows that stopwords are
much stronger hubs than other hubs which we gain with the exclusion of stopwords.

\begin{table}[h!]
\tabcolsep=0.11cm
\caption{Top ten hubs in networks constructed from C3.}
\begin{center}
\begin{tabular}{||c|c|c|c|c|c|c|c||}
\hline
\multicolumn{4}{||c|}{SW included} & \multicolumn{4}{|c||}{SW excluded} \\ \hline
\multicolumn{2}{||c|}{$m_{2}$} & \multicolumn{2}{|c|}{$m_{6}$} & \multicolumn{2}{|c|}{$m_{2}$} &
\multicolumn{2}{|c||}{$m_{6}$} \\ \hline
word & degree & word & degree & word & degree & word & degree \\ \hline
i (and) & 29762 & i (and) & 67890 & kad (when) & 4260 & kad (when) & 14921\\
je (is) & 13924 & je (is) & 53484 & rekao (said) & 2036 & rekao (said) & 5755\\
u (in) & 13116 & se (self) & 42563 & sad (now) & 1494 & jedan (one) & 5142\\
se (self) & 11033 & u (in) & 41188 & re\v{c}e (said) & 1319 & sad (now) & 5062\\
na (on) & 9084 & da (yes, that) & 35632 & jedan (one) & 1318 & ljudi (people) & 4836\\
da (yes) & 8103 & na (on) & 29417 & ima (has) & 1281 & dana (day) & 4679\\
a (but) & 6637 & su (are) & 22366 & ljudi (people) & 1264 & ima (has) & 4406\\
kao (as) & 5452 & a (but) & 21919 & dobro (good) & 1119 & re\v{c}e (said) & 4178\\
od (from) & 4773 & kao (as) & 18141 & dana (day) & 998 & dobro (good) & 3964\\
za (for) & 4708 & ne (no) & 16211 & re\'ci (say) & 968 & \v{c}ovjek (human) & 3496\\ \hline
\end{tabular}
\end{center}
\end{table}

\begin{figure}[h!]
\centering 
\includegraphics[width=0.9\textwidth]{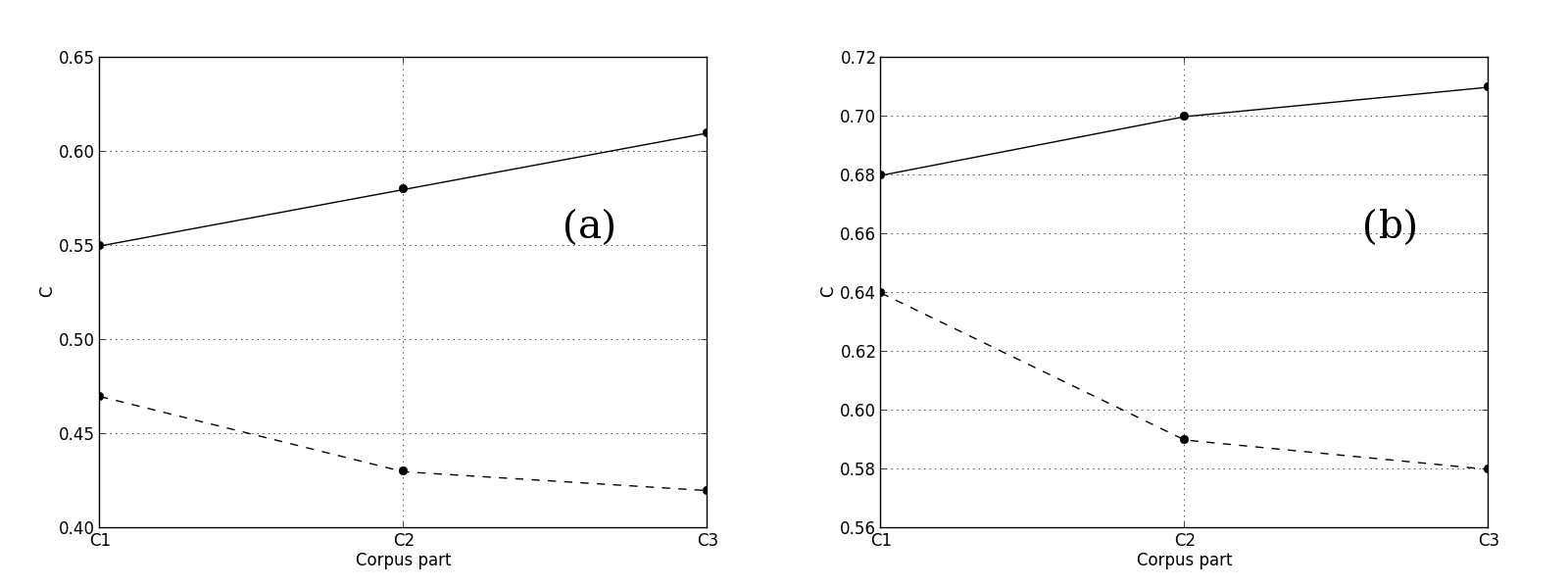}
\caption{The impact of stopwords on the average clustering coefficient in accordance with the various sizes of the corpus parts. $C_{sw}$ (from networks constructed with stopwords included) is represented by solid
lines, while the $C$ (from networks constructed with stopwords excluded) is represented by dashed lines.
(a) $m_{3}$ networks, (b) $m_{6}$ networks.}
\end{figure}

Numerical results of power law distribution analysis indicate the presence of the power law distribution.
The visualization of power law distribution for 4 networks created from C3 is shown in Fig. 3. We
found that networks constructed with included stopwords generally represent a good power law fit
starting from the optimal $x_{min}$. The numeric values of $\alpha$ for the power law distributions
shown in Fig. 2 are respectively: 2.167, 2.172, 2.339, 2.040. The networks with stopwords included have a better power law fit.

\begin{figure}[h]
\centering 
\includegraphics[width=0.9\textwidth]{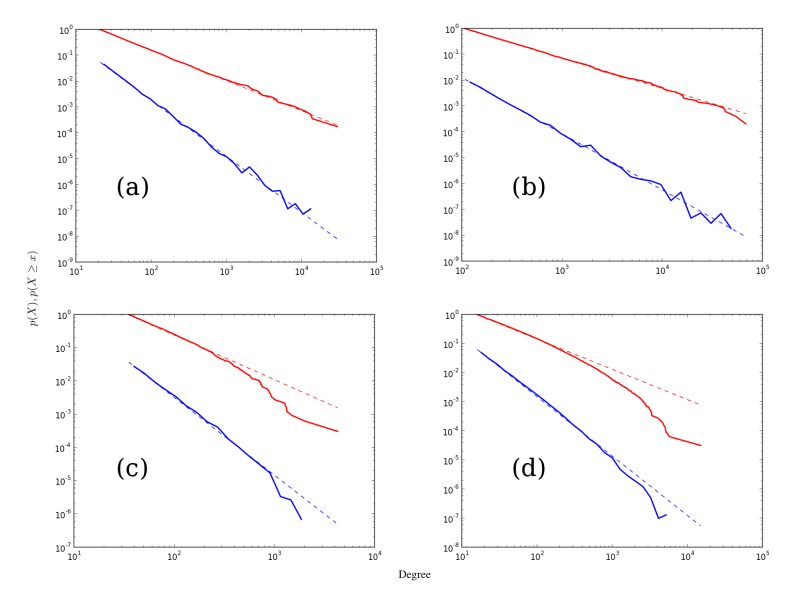}
\caption{Comparison of plots. Probability density function ($p(X)$, lower line) and complementary
cumulative distribution function ($p(X \geq x)$, upper line) of node degrees from networks constructed from
C3: (a) $m_{2}$, stopwords included, (b) $m_{6}$, stopwords included, (c) $m_{2}$, stopwords
excluded, (d) $m_{6}$, stopwords excluded.}
\end{figure}

\section{Conclusion}
In this work we have presented multiple metrics of complex networks constructed as co-occurrence networks
from the Croatian language. Since, the sensitivity of the linguistic network parameters to the corpus size and
stopwords \cite{caldeira2006network, choudhury2010global} is a known problem in the construction of linguistic
networks, we analyzed the Croatian co-occurrence network. We presented the results of 30
networks constructed with the aim to examine variations among: corpus size, stopword removal and the
size of the co-occurrence window. 

The results in Tables 2, 3, 4, are pointing that the increase of the co-occurrence window size is
followed by the diameter $D$ decrease, average path $L$ shortening and expectedly condensing the
average clustering coefficient $C$. It is worth noticing, that the increased window size contributed to
the results the same as the increase of the used quantity of texts did, suggesting emphasized small-world
properties. The larger size of co-occurrence window plays a key role in the strengthening of
properties of the small-world networks. This observation should be considered in detail in the prospect work.  

Furthermore, the inclusion of stopwords in the process of network construction causes the same
effect. It is evident from Table 5 that stopwords, although they have no strong semantic properties,
act as hubs which can be cumbersome for semantic text analysis. The inclusion of stopwords in co-occurrence networks seems to contribute to the benefit of power law distribution, regardless of the
co-occurrence window size. We point out the varying behaviour of the clustering coefficient (dynamics) by increasing the corpus size. According to our results, it depends on the presence of stopwords in the
corpus: increasing the corpus size with stopwords included, increases the value of $C$,
while increasing the corpus size with the stopwords excluded, decreases the value of $C$.

Finally, since the size of texts is reflected in the network properties, our results suggest
that the influence of the corpus can be reduced by increasing the co-occurrence window size. 
This paper is a preliminary study of the Croatian linguistic network, and more detailed research should be performed in the future. Firstly, the results should be tested on a larger corpus and power law and scale
free properties proven. Additionally, the research towards extracting network semantics is a new
and thrilling branch of our pursuit.


\begin{thebibliography}{1}
\bibitem{alstott2013powerlaw}
Jeff Alstott, Ed Bullmore, and Dietmar Plenz. powerlaw: a python package for
analysis of heavy-tailed distributions. \textit{arXiv preprint arXiv:1305.0215}, 2013.

\bibitem{biemann2012quantifying}
Chris Biemann, Stefanie Roos, and Karsten Weihe. Quantifying semantics using complex network analysis. In \textit{COLING}, pages 263-278, 2012.

\bibitem{borge2010semantic}
Javier Borge-Holthoefer and Alex Arenas. Semantic networks: Structure and dynamics. \textit{Entropy}, 12(5):1264-1302, 2010.

\bibitem{caldeira2006network}
Silvia MG Caldeira, TC Petit Lobao, Roberto Fernandes Silva Andrade, Alexis
Neme, and JG Vivas Miranda. The network of concepts in written texts. \textit{The European Physical Journal B-Condensed Matter and Complex Systems}, 49(4):523-529,
2006.

\bibitem{choudhury2010global}
Monojit Choudhury, Diptesh Chatterjee, and Animesh Mukherjee. Global topology
of word co-occurrence networks: Beyond the two-regime power-law. In \textit{Proceedings of the 23rd International Conference on Computational Linguistics: Posters},
pages 162-170. Association for Computational Linguistics, 2010.

\bibitem{dorogovtsev2001language}
Sergey N Dorogovtsev and Jos\'e Fernando F Mendes. Language as an evolving word
web. \textit{Proceedings of the Royal Society of London. Series B: Biological Sciences},
268(1485):2603-2606, 2001.

\bibitem{hagberg2008exploring}
Aric Hagberg, Pieter Swart, and Daniel S Chult. Exploring network structure, dynamics, and function using networkx. Technical report, Los Alamos National Lab-
oratory (LANL), 2008.

\bibitem{i2001small}
Ramon Ferrer i Cancho and Richard V Sol\'e. The small world of human lan-
guage. \textit{Proceedings of the Royal Society of London. Series B: Biological Sciences},
268(1482):2261-2265, 2001.

\bibitem{liu2013language}
HaiTao Liu and Jin Cong. Language clustering with word co-occurrence networks
based on parallel texts. \textit{Chinese Science Bulletin}, 58(10):1139-1144, 2013.

\bibitem{masucci2006network}
AP Masucci and GJ Rodgers. Network properties of written human language. \textit{Physical Review E}, 74(2):026102, 2006.

\bibitem{pardo2006using}
Thiago Alexandre Salgueiro Pardo, Lucas Antiqueira, M das Gracas Nunes,
ON Oliveira, and Luciano da Fontoura Costa. Using complex networks for language
processing: The case of summary evaluation. In \textit{Communications, Circuits and Systems Proceedings, 2006 International Conference on}, volume 4, pages 2678-2682.
IEEE, 2006.

\end{thebibliography}
\end{document}